\documentclass{article} % For LaTeX2e
\usepackage[usenames,dvipsnames]{color}
\usepackage{nips,times}
\usepackage{hyperref}
\usepackage{url}

\usepackage{amsmath}
\usepackage{footnote}
\usepackage{amsthm}
\usepackage{amsfonts}
\usepackage{graphicx}
\usepackage{array}
\makeatletter
\newif\if@restonecol
\makeatother

\usepackage[ruled,vlined,linesnumbered]{algorithm2e}
\usepackage{cite}
\usepackage{mdwlist}
\usepackage{placeins}

\usepackage{mfirstuc}
\usepackage{glossaries}
\usepackage{makeidx}\makeindex

\usepackage{xr}
\newcommand{\mcrot}[4]{\multicolumn{#1}{#2}{\rlap{\rotatebox{#3}{#4}~}}}

\def\accuracy{power}
\def\accuracytitle{Statistical power}
\def\consistency{theoretical asymptotic consistency}

\def\fprboundlong{theoretically derived NFP bound}
\def\fprbound{NFP bound}
\def\fpr{NFP}
\def\fprfirst{number of false positives (NFP)}
\def\fprpure{number of false positives}
\def\tpr{NTP}
\def\tprfirst{number of true positives (NTP)}

\def\problemselect{selection problem}
\def\problemselectP{selection problems}

\def\problemvar{variable selection}

\def\problemvarprobP{variable selection problems}

\def\algoselect{selection algorithm}
\def\algoselectF{Selection algorithm}
\def\algoselectP{selection algorithms}

\def\algoselectPtitle{Selection algorithms}
\def\algostructP{structured selection algorithms}
\def\algostructregP{structured regression algorithms}
\def\structreg{structured regression}
\def\algoregP{regression algorithms}

\def\algovarF{Variable selection algorithm}

\def\algovarP{variable selection algorithms}

\def\modelsizeB{model size}

\def\stabilityselectionfirst{Stability Selection}
\def\stabilityselection{stability selection}
\def\stabilityselectiontitle{Stability selection}
\def\stabilityselectionF{Stability selection}
\def\screenandclean{screen and clean}
\def\screenandcleanF{Screen and clean}
\def\subalgo{sub-algorithm}
\def\subalgoP{sub-algorithms}
\def\lasso{lasso}
\def\lassoF{Lasso}
\def\grouplasso{group lasso}
\def\grouplassoF{Group lasso}
\def\siolfirst{structured input-output lasso (SIOL)}
\def\siolfirstF{Structured input-output lasso (SIOL)}
\def\siolpureF{Structured input-output lasso}
\def\siolpure{structured input-output lasso}
\def\siol{SIOL}
\def\siolF{SIOL}
\def\pursuitfirst{orthogonal matching pursuit (OMP)}
\def\pursuit{OMP}
\def\xval{cross-validation}
\def\xvalF{Cross-validation}

\def\selectableP{elements}

\def\component{component}
\def\componentP{components} 
\def\selectionset{selection set}
\def\dataset{data set}
\def\datasetP{data sets}

\def\datasetPtitle{Data sets}

\def\synthetic{synthetic}
\def\syntheticadverb{synthetically}
\def\hybrid{partially synthetic}
\def\real{real-world}

\def\analyst{analyst}
\def\tuning{tuning parameter}
\def\tuningP{tuning parameters}

\def\bsymb{\mathcal{\hat{B}}}
\def\assumption{assumption}
\def\assumptionP{assumptions}

\title{Stability Selection for Structured Variable Selection}

\author{
George Philipp\\
Carnegie Mellon University\\
Pittsburgh, PA 15213 \\
\texttt{george.philipp@email.de} \\
\And
Seunghak Lee \\
Carnegie Mellon University\\
Pittsburgh, PA 15213 \\
\texttt{leeseunghak@gmail.com} \\
\And
Eric P. Xing \\
Carnegie Mellon University\\
Pittsburgh, PA 15213 \\
\texttt{epxing@cs.cmu.edu} \\
}
\nipsfinalcopy

\begin{document}

\maketitle

\begin{abstract}

In variable or graph selection problems, finding a right-sized model or controlling the number of false positives is notoriously difficult. Recently, a meta-algorithm called \stabilityselectionfirst \ was proposed that can provide reliable finite-sample control of the \fprpure. Its benefits were demonstrated when used in conjunction with the \lasso \ and orthogonal matching pursuit algorithms.

In this paper, we investigate the applicability of \stabilityselection \ to \algostructP: the \grouplasso \ and the \siolpure. We find that using \stabilityselection \ often increases the \accuracy \ of both algorithms, but that the presence of complex structure reduces the reliability of error control under \stabilityselection. We give strategies for setting \tuningP \ to obtain a good model size under \stabilityselection, and highlight its strengths and weaknesses compared to competing methods \screenandclean \ and \xval. We give guidelines about when to use which error control method.

%SS was proposed to improve power, theoretical robustness, reliable error control
%use it on SMR
%R1:consistently more power
%R2:error bound reliable
%R3:tradeoff

\end{abstract}

%\category{I.5.0}{Pattern Recognition}{General}[]
%\keywords{variable selection, subset selection, structure learning, inference, association analysis, association mapping}

\section{Introduction}

Selecting a discrete set of \selectableP \ such as variables or edges in a graphical model is an important problem in modern statistics, arising in key areas such as genome-wide association studies. Oftentimes, there is no ground truth available for such problems and an interpretation of the result is difficult. Most \algoselectP \ allow the \analyst \ to control the number of selected \selectableP \ ({\it \modelsizeB}) by varying \tuningP \ which determine, for example, the amount of regularization used. No matter what values are chosen for these \tuningP, it is often unclear whether too many, too few or just the right number of \selectableP \ have been selected.

% For example, selecting 10 genetic markers as being associated with a trait takes on very different meanings depending on whether the actual number of associated markers is, say, 1 or 100. In general, increasing the \modelsizeB \ raises both the \tprfirst \ and the \fprfirst. The \analyst \ needs to find a good trade-off between these quantities without knowing either. Alternatively, he / she might want to ensure that the \fpr \ does not exceed a certain value, as \fpr \ is often viewed as somewhat more important than \tpr.

A number of solutions have been proposed to address this problem, the main ones being information criteria such as the Akaike Information Criterion (AIC) \cite{AIC} and the Bayesian Information Criterion (BIC) \cite{BIC}, and \xval \ \cite{xVal}. However, none of these perform consistently well in the especially challenging case where the number of \selectableP \ to select from greatly exceeds the number of available data samples \cite{stars}.

Recently, a meta-algorithm called \stabilityselectionfirst \ \cite{ss, stars} was introduced to address this issue. \stabilityselectionF \ can be used in conjunction with any other \algoselect \ (which we call the {\it \subalgo}), and can augment and improve this sub-algorithm in several respects. For sub-algorithms \lasso \ and \pursuitfirst, Meinshausen and B{\"u}hlmann \cite{ss} demonstrated: (1) Under (strong) assumptions, we have a theoretically derived bound on the expected \fprfirst \ when using \stabilityselection. (2) This bound holds up in experiments even under much weaker conditions. (3) \stabilityselectionF \ increases statistical \accuracy. And, (4) \consistency \ of selection holds under more general conditions when \stabilityselection \ is used.  

%\stabilityselectionF \ employs sub-sampling to run the \subalgo \ many times on different, randomly drawn, subsets of the original \dataset. Then, a consensus vote is taken. \selectablePF \ that were selected many times during this procedure enter the final selection. 

%Several results suggest that \stabilityselection \ comes with significant theoretical and practical benefits. 

It is clear that in scenarios where all of these benefits apply, \stabilityselection \ is a powerful tool. Hence, it is important to better understand when they apply. If we can demonstrate that \stabilityselection \ works well when used with more complex \subalgoP, we would significantly broaden its scope of applicability and increase the attractiveness of those \subalgoP. This is the aim of this paper. Specifically, we are interested in a family of recently introduced algorithms called {\it sparse structured multivariate and multi-task regression}, where various forms of structural information in the data is used to boost \accuracy. High performance and practical usability of \algostructregP \ have been demonstrated in various applications \cite{kim2009multivariate, lee2009learning, lee2012leveraging, kim2010tree, jenatton2010proximal, szabo2011online, bazerque2011group, varoquaux2010brain, yang2011tag, lu2010supervised, ma2007supervised}.

%An $L_q$-regularizer is employed as a convex relaxation of the sparsity-maximizing $L_0$-penalty or constraint.

In this paper, we study the \lasso \cite{lasso}, the \grouplasso \ \cite{gl} and the \siolfirst \ \cite{SIOL}. We begin by defining notation and the error control methods studied in this paper in section \ref{definitions}. In section \ref{sserror}, we discuss the applicability of the \fprboundlong \ from \cite{ss}. Then we perform an extensive empirical study on a variety of \synthetic \ and \hybrid \ \datasetP \ for which the ground truth selection is available. We describe the \datasetP \ used in section \ref{datasets} and the results in section \ref{results}. 

The main contributions from our analysis are:

\begin{itemize}
\item We demonstrate that \stabilityselection \ increases the power of \lasso, \grouplasso \ and \siol \ in many settings.
\item We show how the reliability of the \fprboundlong \ for \stabilityselection \ depends on the complexity of the structural information used.
\item We introduce a strategy for automatic calibration of \tuningP \ to achieve a desirable trade-off between \fpr \ and \tprfirst.
\item We highlight the strengths and weaknesses of \stabilityselection \ compared to two competing error control methods, \xval \ and \screenandclean \cite{screenAndClean}, in the context of \structreg \ and give guidelines on when to use which method.
\end{itemize}

We conclude in section \ref{conclusion}.

\section{Definitions and Background} \label{definitions}

We frame a general \problemselect \ as the task of choosing some subset of {\it \selectableP} \ $S$ of a discrete {\it \selectionset} $K$ on the basis of a \dataset \ $X$ with $N$ data points. Our goal is that the chosen $S$ is close to the ``true subset'' $S^*$ (which may not be well-defined in practice). We call its compliment $N^*$ ($ = K \backslash S^*$), the number of false positives $V = |N^* \cap S|$ and the number of true positives $T = |S^* \cap S|$.  A \algoselect \ is a function $f$ that maps $X$ to $S$. %All selection algorithms used in this paper are defined in detail in section S1 of the supplementary information.

All \algoselect s we study in this paper (\lasso, \grouplasso \ and \siol) are for {\it \problemvar}, a specific type of \problemselect. In that setting, we wish to discover statistical associations between $d$ input variables and $t$ output variables (also called {\it tasks}). We denote input variables by integers $\{1,..,d\}$ and the output variables by integers $\{1,..,t\}$. Every data point has $d+t$ dimensions, each corresponding to the value of an input or output variable. The selection set $K$ has $d*t$ elements, each corresponding to a possible association between a single input variable and a single output variable (input-output pair). The three \algoselect s are defined in section \ref{algodetails} in the appendix.

\subsection{\stabilityselectiontitle} \label{ss}

\stabilityselectionF \ is a meta-algorithm that can be used in conjunction with any \algoselect \ $f$ as defined above, and ``$f$ with \stabilityselection'' can itself be viewed as a \algoselect. \stabilityselectionF \ samples a random subset of fixed size $\lfloor pN\rfloor$ from the \dataset \ without replacement. Then it runs $f$ on this subset and obtains a selection $S_1$. This is repeated $I$ times with independently drawn subsets to obtain selections $S_1$, $S_2$, .., $S_I$. The {\it stability} $\tau$ of each element of $K$ is the proportion of these $I$ selections it appears in. The final selection $S^{stable}$ will be the set of elements in $K$ that have stability greater or equal to some threshold $\pi$. \stabilityselectionF \ is shown in algorithm \ref{ssalgo}. 

\begin{algorithm}
 \KwData{\algoselectF \ $f$, choice of \tuningP \ for $f$, \dataset \ $X$ of size $N$, sub-sampling parameter $p$, threshold parameter $\pi$}
 \KwResult{Selection $S^{stable}$}
 $\tau_{1,..,K} = 0$ \;
 \For{$iter = 1$ to $I$}{
  Select random subset $X_{iter}$ of \dataset \ $X$ of size $\lfloor pN\rfloor$\;
  Run $f$ on $X_{iter}$ with chosen \tuningP \ to obtain selection $S_{iter}$\;
  \ForAll{$k$ in $S_{iter}$}{
   $\tau_k = \tau_k + \frac{1}{I}$\;
  }
 }
 $S^{stable} = \{k \in K : \tau_k \ge \pi\}$
 \caption{\stabilityselectionF : %basic form
 }
\caption{\stabilityselectionF}
\label{ssalgo}
\end{algorithm}

Meinshausen and B{\"u}hlmann \cite{ss} used a slightly different form of \stabilityselection. They considered the maximum stability over different values of $\lambda$. We did not follow this practice because it makes the algorithm more complicated, less universal (the sub-algorithm must have a $\lambda$-parameter) and yielded, on average, worse results in our experiments.

\subsection{\xvalF \ for \algoselectP} \label{crossval}

A classic approach to setting \tuningP \ in many selection settings \ is $k$-fold \xval. Under this procedure, the dataset is partitioned into $k$ {\it folds} of equal or near-equal size. One of those folds is then considered the {\it validation set} and the remaining folds together the {\it training set}. For a given \tuning \ configuration, the following three steps are performed.

%For a given training and validation set, and a given \tuning \ configuration, we follow the following steps:

\begin{enumerate}
\item Run the \algoselect \ on the training set to obtain an initial selection.
\item Fit a predictive model on the validation set whose complexity is controlled by the initial selection.
\item Calculate the error of the predictive model on the validation set. This is the {\it validation error}.
\end{enumerate}

These steps are then repeated for all considered \tuning \ configurations and all $k$ possible validation sets. The \tuning \ that achieves the least mean validation error across the $k$ validation sets is ultimately chosen. The final selection is then the selection obtained when running the \algoselect \ on the whole dataset.

For \problemvarprobP, the predictive model from step 2 is usually (unregularized) least squares regression from the input to the output variables where all regression coefficients corresponding to input-output pairs not in the initial selection are fixed to be zero. This is the choice we follow in this paper.

\subsection{\screenandcleanF} \label{sac}

\screenandcleanF, like \stabilityselection, is a meta-algorithm that augments \algoselectP \ to enable error control. \screenandcleanF \ consists of a {\it screen phase} and a {\it clean phase}. In the screen phase, promising elements of the \selectionset \ are being singled out using \xval. \screenandcleanF \ makes use of the tendency of \xval \ to over-select. This tendency ensures that most positives enter the clean phase, where we hope to remove many remaining false positives.

\begin{algorithm}
 \KwData{\algovarF \ $f$, \dataset $X$, threshold $\pi$}
 \KwResult{Selection $S^{SaC}$}
 \For{$iter = 1$ to $I$}{
  Randomly split $X$ into two shards $X^1_{iter}$ and $X^2_{iter}$ of equal size\;
  Obtain $S_{iter}$ by running $f$ on $X^1_{iter}$, setting \tuningP \ for $f$ by $k$-fold \xval\;
  \For{$j=1$ to $t$}{
   $S_{iter}(j) = $ all $i$ with $(i,j) \in S_{iter}$\;
   Obtain the linear regression model where output $j$ is the response, all inputs $i$ with $i \in S_{iter}(j)$ are predictors, and the \dataset \ used is $X_{iter}^2$\;
   \ForAll{$i \in S_{iter}(j)$}{
    $p_{(i,j)} = $ p-value for the t-statistic obtained from the linear model for input $i$\;
    $p^{array}_{(i,j)}[iter] = p_{(i,j)} * |S_{iter}(j)| * t$\;
   }
   \For{$i = 1$ to $d$, $i \not \in S_{iter}(j)$}{
    $p^{array}_{(i,j)}[iter] =  d * t$\;
   }
  }
 }
 \For{$j=1$ to $t$}{
  \For{$i=1$ to $d$}{
   $p_{(i,j)} = FDR(p^{array}_{(i,j)})$\;
  }
 }
 $S^{SaC} = \{(i,j) \in K : p_{(i,j)} \ge \pi\}$
\caption{\screenandcleanF \ for \problemvar}
\label{sacalgo}
\end{algorithm}

We sketch \screenandclean \ for \problemvar \ in algorithm \ref{sacalgo}. $FDR$ stands for the Benjamini-Hochberg False Discovery Rate \cite{pvals}. \screenandcleanF \ produces p-values for each input-output pair. As the name suggests, admitting all pairs with p-value less than $\pi$ ($\pi < 1$) into the final selection bounds the probability of obtaining a false positive at $\pi$, as long as the assumptions that were used to derive the p-values hold. Just as the \fprbound \ for \stabilityselection, these conditions do not hold exactly in practice and so the reliability of the p-values varies. 

We have deliberately phrased \screenandclean \ in algorithm \ref{sacalgo} to allow pseudo-p-values larger than 1. Thus, we can meaningfully set $\pi$ to values greater than 1 and still bound the expected number of false positives at that value. FDR can be used to combine such pseudo-p-values. 

The central idea behind \screenandclean \ is that the screen phase reduces the number of variables considered for selection in the clean phase, which reduces the magnitude of the correction needed for multiple testing (line 9 in algorithm \ref{sacalgo}).

\section{Error control with \stabilityselection} \label{sserror}

Meinshausen and B{\"u}hlmann \cite{ss} proved a theoretical bound on the expected \fpr \ when using \stabilityselection, under strong \assumptionP \ that are unachievable in practice. In experiments with the \lasso, however, this bound turned out to be quite reliable. To predict when it is reliable, we must understand how closely the \assumptionP \ made resemble reality. 

Fix $N$ and consider the following \assumptionP: (1) Data points are drawn independently from some distribution $P$. (2) There is a constant $c$ such that for all $k\in N^*$ we have $\mathbb{P}(k \in S_{\lfloor \frac{1}{2}N\rfloor}) = c$, where $S_{\lfloor \frac{1}{2}N\rfloor}$ is the selection of the \subalgo \ on a random \dataset \ of size $\lfloor \frac{1}{2}N\rfloor$ drawn from $P$. (3) The \subalgo \ is not worse than random guessing, i.e. $\frac{\mathbb{E}(|S^* \cap S_{\lfloor \frac{1}{2}N\rfloor}|)}{\mathbb{E}(|N^* \cap S_{\lfloor \frac{1}{2}N\rfloor}|)} \ge \frac{|S^*|}{|N^*|}$. Let $q = \mathbb{E}(|S_{\lfloor \frac{1}{2}N\rfloor}|)$. Under these \assumptionP, the number of false positives of \stabilityselection \ with sub-sampling parameter $p = \frac{1}{2}$ and selection threshold $\pi$ is bounded, in expectation over random \datasetP \ of size $N$, by:

\begin{equation} \label{eqbound}
\mathbb{E}(V^{stable}) \le \frac{q^2}{|K|(2\pi - 1)}
\end{equation}

We define $\mathcal{B}$ to be $\frac{q^2}{|K|(2\pi - 1)}$.

The strongest \assumption \ is (2). It says that each mistake is equally likely, which implies that running the \subalgo \ on independent data sets will likely produce different mistakes. \stabilityselectionF \ is based on running the \subalgo \ multiple times on partially independent \datasetP \ for validation. Assumption (2) ensures the effectiveness of this. We will discuss how realistic this assumption is in the case of {\it \problemvar}, an important class of \problemselectP \ that we study in this paper. %In \problemvar \, we wish to discover statistical associations between $d$ input variables and $t$ output variables (also called {\it tasks}). We denote input variables by integers $\{1,..,d\}$ and the output variables by integers $\{1,..,t\}$. Every data point has $d+t$ dimensions, each corresponding to the value of an input or output variable. $K$ has $d*t$ elements, each corresponding to a possible association between a single input variable and a single output variable.

For \algovarP \ that are symmetric in the input variables such as \lasso \ and \pursuit, condition (2) is fulfilled, for example, when (a) the input variables are exchangeable and (b) the output variables are independent of the input variables they are not associated with when conditioned on the input variables they are associated with. An example of an exchangeable distribution is a joint Gaussian distribution where each diagonal element of the covariance matrix has the same value and each off-diagonal element has the same value. However, any small perturbation of the covariance matrix would cause \assumption \ (2) to be violated. In practice, of course, certain pairs of predictors tend to be more correlated than others, as in the case of linkage disequilibrium in genome-wide association studies. This may violate condition (2). 

When we introduce structure such as group structure, the \algoselect \ is no longer symmetric with respect to all input variables. This may make (2) more unrealistic by introducing asymmetry in the selection probabilities of the negatives. We can ask three questions to gauge the size of this effect: (i) Are some groups partially contained in $S^*$? (ii) How symmetric is the group structure? And, (iii) what is the weight of the group-based penalties compared to individual-component penalties?

If a group is partially contained in $S^*$, the negatives in that group tend to be more likely to be selected than negatives that do not share a group with one or more positives. For example, if groups do not overlap and we know / assume that $S^*$ is a union of groups, then the accuracy of the \fprbound \ should not be compromised as much as if groups overlap or we wish to discover isolated positives within groups. 

If the group structure is asymmetric, it is likely that some negatives will be favored over others. For example, if we have group information on certain input variables but not on others, the total amount of regularization acting on each variable may differ between these two types of input variables. Similarly, groups of different size may behave differently. In general, larger groups of negatives are more likely to be selected then smaller ones. (This can be mitigated to some degree by giving larger groups a higher weight.) Similarly, variables in the overlap of two groups are treated differently from variables that are exclusively in one group. Also, if the group penalties overall have a high weight compared to the individual-component penalties (or there are no individual-component penalties), the problems described may be exacerbated.

For example, \assumption \ (2) holds when the input distribution is exchangeable, the groups are disjoint, of equal size and cover all input variables, and $|S^*|$ does not contain partial groups. If the groups were of unequal size, we would have to weight each group penalty by the square root of its size and assume that the input variables are pairwise independent to maintain (2).

\paragraph{Summary} We expect error control under \stabilityselection \ to work better when there are no groups, the group structure is simple or the weight on the group penalties is small compared to individual-component penalties. We will investigate this hypothesis in section \ref{results}.

Note that the \fprbound \ (\ref{eqbound}) uses the quantity $q$, which is an expectation over a randomly drawn \dataset. For \real \ \datasetP, of course, there is no ``distribution'' from which the \dataset \ is drawn. Hence, we must use the empirical estimate $\hat{q} = \frac{\sum_{i=1}^I |S_i|}{I} = \sum_k \tau_k$ (notation is the same as in algorithm \ref{ssalgo}). While $q$ could be estimated accurately in experiments with \synthetic \ data where we control the data distribution $P$, we will use $\hat{q}$ in all our experiments to make them as realistic as possible. We conjecture that $V - \frac{\hat{q}^2}{|K|(2\pi - 1)}$ may actually have a lower variance than $V - \frac{q^2}{|K|(2\pi - 1)}$ in practice. We denote $\frac{q^2}{|K|(2\pi - 1)}$ by $\mathcal{B}$ and $\frac{\hat{q}^2}{|K|(2\pi - 1)}$ by $\bsymb$.

\subsection{Parameter tuning} \label{tuning}

As we vary the \tuningP \ of any \algoselect, we will obtain varying numbers of true positives and false positives. It is reasonable to assume that we can express our preference for a \tuning \ setting as an objective function $v(.)$ that depends on $T$ and $V$. Our goal is then to find \tuningP \ such that $v(T,V)$ is maximized. Of course, $T$ and $V$ are unknown. However, we may use the  \fprbound \ and proxy $V$ with $\bsymb$ and consequently $T$ with $|S| - \bsymb$. Hence, our goal becomes to maximize $v(|S| - \bsymb, \bsymb)$. We will investigate this strategy in section \ref{choice}.

\section{\datasetPtitle} \label{datasets}

In this section, we describe the \datasetP \ we generated for our empirical study. In order to evaluate \algoselectP, it is necessary to use data with available ground truth selection $S^*$. In the \problemvar \ setting, this is conventionally achieved by \syntheticadverb \ generating output variables as functions of several input variables, plus noise. A pair of an input and an output variable is contained in $S^*$ if the input variable was used to generate the output variable, and then we say the variables are {\it associated}. %Output variables generated this way are independent of the input variables they are not associated to when conditioned on the input variables they are associated to. 

%The input variables can stem from a ``\real'' \dataset \ or be \synthetic \ also. In the first case, we call the resulting \dataset \ {\it \hybrid}. The full set of input variables is also called {\it input vector} and the set of all input vectors in the \dataset \ the {\it design matrix}. 

In this paper, we use seven {\it types} of design matrices. The first five types (A-E) are \synthetic. They correspond to a particular recipe for drawing a (random) matrix. Type F is also \synthetic \ but refers to a specific fixed \dataset. Type G is a specific \real \ \dataset \ and is also fixed. The seven types are very similar to those used in \cite{ss} and are as follows.

\begin{itemize}
\item [{\bf A}] Input vectors are drawn from $\mathcal{N}(0,I)$.
\item [{\bf B}] Input vectors are drawn from $\mathcal{N}(0,\Sigma)$, where $\Sigma$ is block-diagonal with 10 blocks of equal size. $\Sigma_{ij}$ is 1 if $i=j$, 0.5 if $i$ and $j$ are in the same block and 0 otherwise.
\item [{\bf C}] Input vectors are drawn from $\mathcal{N}(0,\Sigma)$, where $\Sigma_{ij} = 0.99^{|i-j|}$. Hence, $\Sigma$ is Toeplitz.
\item [{\bf D}] Input vectors are drawn from a factor model. We first draw two factors $f_1$ and $f_2$ according to $\mathcal{N}(0,I)$ distributions for the entire \dataset. Then, the input vectors are generated as $af_1 + bf_1 + \epsilon$, where $a$ and $b$ are $\mathcal{N}(0,1)$ scalars and $\epsilon$ is a $\mathcal{N}(0,I)$ vector.
\item [{\bf E}] Same as type D but with 10 factors instead of 2.
\item [{\bf F}] A simulated human genome \dataset \ with 5000 input variables and 1000 data points. This was generated using a software called ``GWAsimulator'' \cite{humandata}. 
\item [{\bf G}] A yeast genome \dataset \ with 1260 input variables and 114 data points \cite{yeastdata}. 
\end{itemize}

To build a \dataset, we first generate a design matrix $X_{in}$ by picking a type and, for types A-E, choosing $d$ and $N$ and performing the random draw. Then, we normalize mean and variance of each input variable and choose a ground truth $S^*$. We generate a matching $\beta^*$ where components corresponding to input-output pairs in $S^*$ are drawn from the uniform distribution on $[0,1]$ and the remaining components are 0. Next, we generate the response matrix $X_{out}$ according to $X_{out} = X_{in}\beta^* + \epsilon$, where $\epsilon$ has independent Gaussian components with variance $\sigma^2$. We choose $\sigma^2$ to (approximately) achieve a certain signal-to-noise ratio (snr) in our \dataset. Specifically, we set $\sigma^2 = \frac{||X\beta^*||_2^2}{t*N*snr}$, where $snr$ is the target snr. Finally, we normalize both mean and variance of the output variables. %This step alters the effective snr, but only by a negligible amount.

We generate different \datasetP \ for \lasso, \grouplasso \ and \siol. For \lasso, we generate a single output variable and $S^*$ contains $s$ uniformly randomly chosen input variables, where $s$ is a parameter we vary. For \grouplasso, we choose groups $g_i = \{4i-3, 4i-2, 4i-1, 4i \}, 1 \le i \le \frac{d}{4}$, and the ground truth $S^*$ is the union of $s_g = \frac{s}{4}$ uniformly randomly chosen groups. For \siol, we choose input group $g_i$ to range from $\max(5*(i-1), 1)$ to $\min(5*i + 1, d)$, where $1 \le i \le \frac{d}{5}$. We generate five output variables in a single output group.  $S^*$ contains all input-output pairs where the input variable is from one of $s_g$ randomly chosen input groups. Hence, each output variable is associated to the same input variables. Note that because groups are overlapping, for a fixed $s_g$, $|S^*|$ varies.

We call a {\it data configuration} a choice of a design matrix type and values of $d$, $N$, $s$ / $s_g$ and $snr$, whenever these parameters are applicable. We generate a total of 5600 data sets each for \lasso \ and \grouplasso \ (100 random draws for each of 56 configurations) and 2600 for \siol \ (100 random draws for each of 26 configurations). The configurations used are shown in tables \ref{lassoconfigs}, \ref{glconfigs} and \ref{siolconfigs}. We choose configurations similar to \cite{ss}.

\section{Empirical Results} \label{results}

In this section, we show and discuss the results of our experiments. In subsection \ref{accuracy}, we investigate the ability of \stabilityselection \ to increase the \accuracy \ of each sub-algorithm, i.e. its ability to raise the ROC curve. In subsection \ref{control}, we investigate the reliability of the \fprbound \ provided for \stabilityselection \ by \cite{ss}. In subsection \ref{choice}, we investigate the possibility of setting \tuningP \ automatically to control the trade-off between \fpr \ and \tpr. In the appendix, we present more detailed results (section \ref{detailedresults}), compare the runtime of algorithms (section \ref{runtime}) and provide more experimental details (section \ref{expdetails}).

For convenience, let us define two terms. From now on, we will use the word {\it algorithm} to refer to one of \{\lasso, \grouplasso, \siol\} and the word {\it regime} to refer to whether we are running an algorithm with Stability Selection, with Screen and Clean or as baseline (= without either of the two).

\subsection{\accuracytitle} \label{accuracy}

In this section, we compare the ROC curves under different regimes. For \screenandclean, the trade-off between \tpr \ and \fpr \ is controlled by a single parameter, the selection threshold $\pi$. Under the baseline regime, it is controlled by the parameter $\lambda$. \stabilityselectionF \ has both of these parameters. Hence, to make the comparison fair, we automatically set $\lambda$ to achieve $\hat{q} \approx \sqrt{0.8*|K|}$ and then only vary the threshold parameter $\pi$ to generate the ROC curve. This is a heuristic introduced in \cite{ss}. 

\begin{figure*}
\centering
\includegraphics[scale=1]{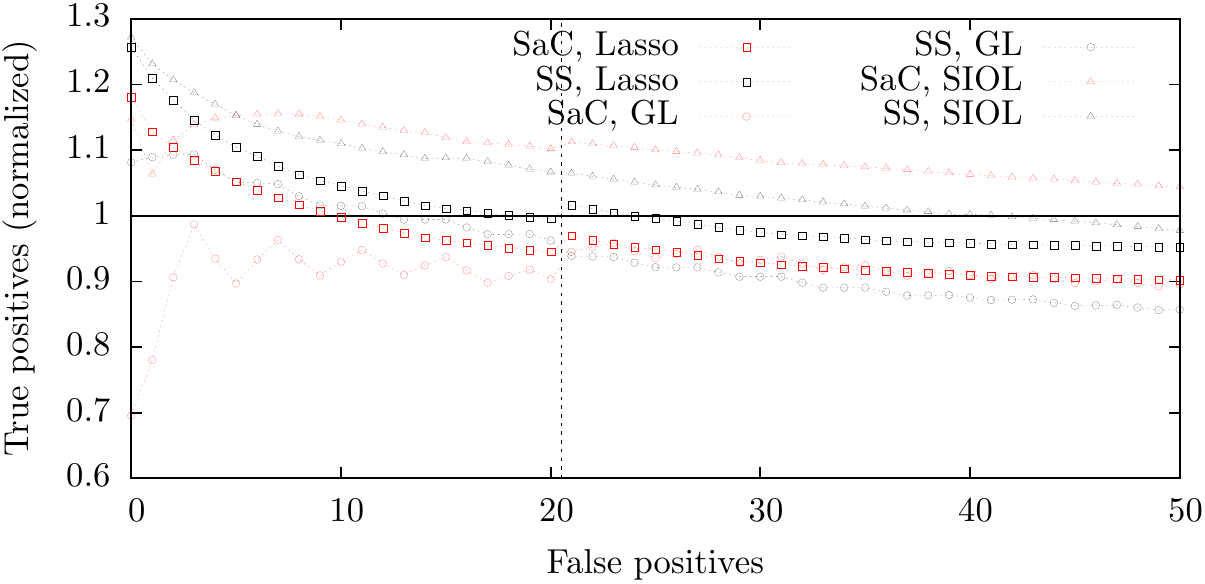}
\caption{ROC curve relative to baseline, by regime and algorithm. Results are averaged over all \datasetP. Beyond 20 false positives, only \datasetP \ with at least 500 data points and input variables are considered.}
\label{roc}
\end{figure*}

In figure \ref{roc}, we show, for each of the three algorithms, the ROC curve of \stabilityselection, normalized by the average of the ROC curves of \stabilityselection \ and the baseline. We show the average of the curves across all of our \datasetP \ (as described in section \ref{datasets}). We also show the equivalent curves for \screenandclean. 

For all algorithms, we see that the \accuracy \ is improved by \stabilityselection \ and \screenandclean \ with respect to the baseline in the critical scenario when the \fpr \ is small, except for \grouplasso \ with \screenandclean. Also, for small \fpr, \stabilityselection \ outperforms \screenandclean. Importantly, \stabilityselection \ performs consistently well across data configurations. When the \fpr \ is 0, \stabilityselection \ outperforms the baseline in 51/56 configurations for \lasso, 42/56 configurations for \grouplasso \ and 21/26 configurations for \siol. There are only four configurations across all algorithms where \stabilityselection \ finds less than 70\% as many false positives as the baseline. 

A breakdown of results by configuration for the case $NFP = 0$ is given in section \ref{detailedresults}. While \stabilityselection \ consistently outperforms the baseline, \screenandclean \ behaves more erratically. Surprisingly, the performance gap between regimes can change significantly between data configurations that are very similar. It is not clear which factors in the data cause each regime to perform well or badly.

As the \fpr \ increases, the performance of \stabilityselection \ and \screenandclean \ decline with respect to the baseline, suggesting it is more advantageous to use these regimes when we are looking to obtain a small model size. \stabilityselectionF \ still outperforms \screenandclean \ for larger values of \fpr \ for \lasso \ and \grouplasso, and underperforms for \siol.

We hypothesize that one reason for the difference in performance between \stabilityselection \ and \screenandclean \ is that \screenandclean \ does not use group information during the clean phase, as it simply performs least squares regression. This is a disadvantage for detecting weak associations of input variables that share a group with other, more strongly associated input variables. Such variables may be discarded in the clean phase, whereas they may be ``saved'' by the group structure under the other regimes. This disadvantage is particularly strong for the \grouplasso, as the penalties are completely group-based. On the other hand, discarding group information in the clean phase is an advantage when it comes to discarding negatives that share a group with one or more positives. This advantage appears if there are groups which are partially contained in $S^*$, as in our setup for \siol.

\subsection{Reliability of the theoretical \fprbound} \label{control}

To study whether the \fpr \ bound for \stabilityselection \ holds in practice, we simply verify whether $\bsymb$ exceeds $V$. Of course, this will not always hold. We only have a bound on the {\it expected} number of false positives, where the expectation is taken over random draws of the \dataset. Also, the \assumptionP \ required for the bound may be more or less unrealistic, depending on the data and algorithm. \\

\begin{figure*}
\centering
\includegraphics[scale=1]{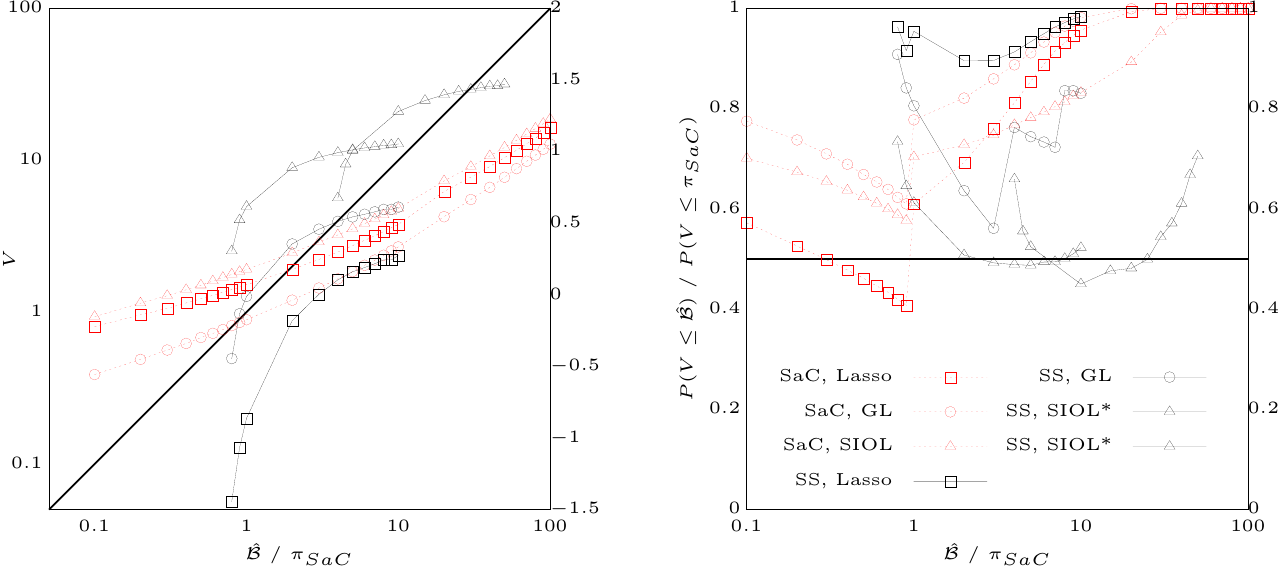}
\caption{Observed number of false positives against theoretical upper bound, by regime and algorithm. On the left, we show the average number of false positives over all \datasetP. On the right, we show the probability that $V$ is below the theoretical upper bound. Notice the use of log scales. For $\pi_{SaC} > 10$, we only consider data sets with at least 500 input variables and data points. For \siol \ with \stabilityselection, we show two curves, one where $\lambda$ is set to achieve $\hat{q} \approx \sqrt{0.8*|K|}$, and one where it is set to achieve $\hat{q} \approx \sqrt{0.8*|K|*t}$.}
\label{errorcontrolcurves}
\end{figure*} 

In section \ref{accuracy}, we showed that overall, \stabilityselection \ has high \accuracy \ when setting $\lambda$ to achieve $\hat{q} \approx \sqrt{0.8|K|}$. We will continue to use this heuristic in this section. In figure \ref{errorcontrolcurves}, we show the average number of false positives $V$ obtained under \stabilityselection \ and the probability of obtaining a number of false positives less than or equal to $\bsymb$. The averages / probabilities are with respect to all \datasetP. The value of $\bsymb$ when varying $\pi$ is shown on the x-axis. We also show equivalent results for \screenandclean, where the value of the selection threshold $\pi_{SaC}$ itself acts as the bound on the expected \fpr, because \screenandclean \ assigns a p-value to each input-output pair.

First, notice that if $\hat{q} \approx \sqrt{0.8|K|}$, $\bsymb$ can only lie, approximately, within the range $[0.8,10]$. This is because $\bsymb = \frac{\hat{q}^2}{|K|(2\pi-1)} \approx \frac{0.8|K|}{|K|(2\pi-1)} = \frac{0.8}{2\pi-1}$. $\pi$ can vary meaningfully between $1$ and just above $0.5$, which leads to the aforementioned range of $\bsymb$.

\begin{table*}
\centering
\begin{tabular}{lllccccccc}
Regime & Algorithm & $\bsymb$ / $\pi_{SaC}$ & A & B & C & D & E & F & G \\ \hline \hline
SaC & \lasso & 1 & 0.07 & 0.66 & {\color{red} 3.55} & 0.76 & {\color{Red} 2.23}& {\color{Red} 2.56} & {\color{Red} 1.11} \\
SS & \lasso & 1 & 0.00 & 0.02 & 0.03 & 0.04 & 0.51 & {\color{Red} 1.04} & 0.05 \\
SaC & \lasso & 10  & 0.67 & 2.82 & 8.24 & 2.31 & 5.20 & 4.07 & 2.73 \\
SS & \lasso & 10 & 0.43 & 1.62 & 1.60 & 1.79 & 4.82 & 4.20 & 1.61 \\
SaC & \grouplasso & 1 & 0.03 & 0.44 & 0.22 & 0.79 & {\color{Red} 2.67} & 0.59 & 0.14 \\
SS & \grouplasso & 1 & 0.00 & 0.07 & 0.13 & 0.99 & {\color{Red} 4.21} & {\color{Red} 1.31} & 0.45 \\
SaC & \grouplasso & 10 & 0.28 & 2.62 & 1.98 & 2.46 & 6.14 & 1.53 & 0.55 \\
SS & \grouplasso & 10 & 0.14 & 2.34 & 3.54 & 4.34 & {\color{Red} 11.70} & 5.32 & 3.16 \\
SaC & \siol & 1 & 0.05 & 0.72 & 0.19 & {\color{Red} 2.70}& {\color{Red} 7.88} & 0.87 & 0.03 \\
SS & \siol & 1 & 0.00 & 0.00 & 0.00 & {\color{Red} 5.93} & {\color{Red} 9.28} & {\color{Red} 12.95} & {\color{Red} 7.98} \\
SaC & \siol & 10 & 0.47 & 3.15 & 1.54 & 6.48 & {\color{Red} 17.11} & 2.31 & 0.32 \\
SS & \siol & 10 & 0.01 & 0.79 & 0.00 & {\color{Red} 18.31} & {\color{Red} 24.05} & {\color{Red} 28.44} & {\color{Red} 23.15}\\
SS* & \siol & 5 & 0.01 & 1.02 & 0.05 & {\color{Red} 11.39} & {\color{Red} 23.96} & {\color{Red} 27.00} & {\color{Red} 23.46}\\
SS* & \siol & 50 & 0.53 & 16.33 & 7.17 & 38.89 & {\color{Red} 59.27} & {\color{Red} 61.41} & 43.89
\end{tabular}
\caption{Average number of false positives observed ($V$) by design matrix type when setting the \fprbound \ to a certain value by tuning $\pi$. For \stabilityselection, we use the heuristic $\hat{q} \approx \sqrt{0.8|K|}$ to set $\lambda$, except rows marked with * use $\hat{q} \approx \sqrt{0.8|K|t}$.}
\label{errorcontroltable}
\end{table*}

Our results are in line with our hypothesis from section \ref{sserror} that more complex structure leads to a decline in the error control capabilities of \stabilityselection. We chose our \datasetP \ / group structure to make \siol \ more challenging than \grouplasso \ under the criteria discussed in section \ref{sserror}. When $\bsymb = 1$, the average number of false positives obtained is 0.20 for \lasso, 1.27 for \grouplasso \ and 4.95 for \siol. This result for \siol \ becomes 5.64 when the group penalty is multiplied by 2 relative to the individual-component penalty and 4.44 in the reverse case. As expected, a relatively larger group penalty leads to a less reliable \fprbound.

Interestingly, even for \siol, the probability of exceeding the bound across all \datasetP \ never falls significantly below 0.5. In other words, the median of $V$ only exceeds $\bsymb$ slightly. This indicates that the distribution of the number of false positives is very sparse across \datasetP. In fact, setting $\bsymb = 1$ yields an average of 12.95 false positives for design matrix type (F) and less than 0.01 false positives for design matrix types (A), (B) and (C). Hence, $\bsymb$ is not only unreliable for \siol, but it may be either a gross underestimate or overestimate depending on the data. The complete breakdown by design matrix type is shown in table \ref{errorcontroltable}.

A possible reason for this volatility is that large blocks of overlapping groups often mean that either no negatives or a large number of negatives are selected. In our experiments with \siol, we often saw that $V$ suddenly jumps from 0 to 10 or 15 as $\pi$ decreases. As we cannot feasibly set $\bsymb$ to 15, this poses a problem for the usability of the \fpr \ bound. For this reason, we also generated results for the alternative heuristic $\hat{q} \approx \sqrt{0.8|K|t}$. This causes $\bsymb$ to lie in the range $[0.8t,10t]$, which is a natural choice for problems with multiple output variables. We show the results in figure \ref{errorcontrolcurves} and table \ref{errorcontroltable} and observe that the problems are only slightly mitigated. For design matrix types (E) and (F), we still have $V >> \bsymb$ and for design matrix types (A), (B) and (C), we still have $V << \bsymb$. This suggests that the pathology is not caused by the choice of $\lambda$.

The accuracy of the \fprbound \ of \screenandclean \ is affected less by structure and also varies less between design matrix types. Hence, \screenandclean \ may be a better choice for error control when complex structure is present or the \dataset \ is challenging. Note that while the threshold $\pi_{SaC}$ can be meaningfully set outside the range $[0.8,10]$ (as opposed to $\bsymb$), figure \ref{errorcontrolcurves} shows that the bound is of poor quality in that region.

\subsection{Model Choice} \label{choice}

In this section, we investigate the usefulness of the \fprbound \ of \stabilityselection \ for automatic parameter tuning as outlined in section \ref{tuning}. We will study the very simple objective function $v(T, V) = T - V$, that can be approximated as $\hat{v}(|S|, \bsymb) = |S| - 2*\bsymb$. While this objective function has shortcomings (in our most challenging \datasetP, it is maximized by the empty selection ...) it is simple and there does not exist a single objective function that is ``right'' in all practical situations.

\begin{savenotes}
\begin{table*}
\centering
{
\footnotesize
\begin{tabular}{lcccccc}
\mcrot{1}{l}{45}{Error Control Scheme} &
\mcrot{1}{l}{45}{Lasso : True Positives} &
\mcrot{1}{l}{45}{Lasso : False Positives} &
\mcrot{1}{l}{45}{GL : True Positives} &
\mcrot{1}{l}{45}{GL : False Positives} &
\mcrot{1}{l}{45}{SIOL : True Positives} &
\mcrot{1}{l}{45}{SIOL : False Positives} \\
SaC \footnote{We maximize over $\pi$: $|S| - 2*\pi$ and $T - V$ (in brackets)} & 3.74 (4.54)& 1.56 (0.38)& 4.41 (12.36)& 1.19 (3.17)& 19.64 (62.06)& 2.97 (19.65) \\
SS (fixed $\lambda$) \footnote{We set $\lambda$ to achieve $\hat{q} \approx \sqrt{0.8*|K|}$ and maximize over $\pi$: $|S| - 2*\bsymb$ and $T - V$ (in brackets)} & 3.10 (4.48)& 0.75 (0.41)& 7.02 (13.55)& 3.84 (2.03)& 19.86 (62.32)& 12.02 (21.78) \\
SS \footnote{We maximize over $\lambda$ and $\pi$: $|S| - 2*\bsymb$ and $T - V$ (in brackets)}& 3.48 (6.46)& 1.59 (0.61)& 10.63 (23.94)& 5.14 (2.45)& 69.12 (122.40)& 57.09 (37.47) \\
Cross-val. \footnote{We simply run \xval \ (no maximization). Values in brackets are obtained from optimizing $T - V$ over $\lambda$ in the baseline regime} & 7.30 (4.21)& 14.00 (0.75)& 20.79 (16.53)& 18.58 (3.39)& 129.41 (82.58)& 144.45 (33.98)
\end{tabular}
}
\caption{Selections obtained from maximizing objective function $v(T,V) = T - V$. We show \tpr \ and \fpr \ obtained from empirical estimates and the ``true optima'' in brackets. Results are averaged over all \datasetP.}
\label{choicetable}
\end{table*}
\end{savenotes}

In table \ref{choicetable}, we show the number of true positives and false positives at the parameter configuration where the objective function is maximized, both using proxies and using the actual values for $V$ and $T$. (The latter are shown in brackets.) We find that both \stabilityselection \ (with $\lambda$ fixed to achieve $\hat{q} \approx \sqrt{0.8|K|}$) and \screenandclean \ find points close to the optimum for \lasso, but their performance degrades for the \grouplasso, and substantially for \siol. This is no surprise given that we showed that with more complex structure, it becomes harder to estimate the number of false positives. Also, we cannot expect to get close to the optimum if the \fpr \ there exceeds 10. One option for mitigating this is to optimize the objective with respect to both $\lambda$ and $\pi$ under \stabilityselection \ (line 3 in table \ref{choicetable}). While optimizing over too many parameters may lead to over-fitting, we do find significantly more positives for \siol.

For all algorithms, we notice that \stabilityselection \ and \screenandclean \ do allow us to find model sizes where the majority of selected variables are true positives. This cannot be achieved by \xval, where in 63\% of all data configurations, more negatives than positives are selected. \screenandcleanF \ seems to be particularly suited for this. For \siol, only 2.97 negatives are selected on average. We may be able to adjust our objective function to encourage small model sizes and achieve an even better ratio between the number of positives and negatives selected. \xvalF, on the other hand, should be used for obtaining large selections that contain many positives and negatives, possibly for further processing (as is done inside the \screenandclean \ algorithm). In this regard, the error control schemes complement each other.

\section{Discussion and Conclusion} \label{conclusion}

In this paper, we studied the interaction of \stabilityselection\  with \algostructP. We found that \stabilityselection \ raises the statistical power in the critical case when there are few false positives, in most of the settings considered. Hence, even in the absence of error control requirements, it may be preferable to run \stabilityselection. If the \algoselect \ uses no structure or the structure is simple, the \fprbound \ can be used to estimate the number of false positives and set \tuningP \ automatically to find a model size where the majority of selected variables are true positives. Such model sizes cannot be found using \xval. However, \xval \ excels at selecting large model sizes that contain the majority of positives.

In settings with complex structure (\siol), the error control mechanisms or \stabilityselection \ are inaccurate and we found \screenandclean \ to be a better alternative that offers very similar benefits.

This paper has two major caveats. First, it depends heavily on heuristics. Our strategy for setting $\lambda$ and our choice of objective function in section \ref{choice} are the most prominent examples. There is no guarantee that some of the limitations found in this paper cannot be overcome by better heuristics. Our goal in this paper was to explore one set of heuristics in depth. We hope this paper can serve as a starting point for exploring other heuristics.

The second caveat is that it is difficult to make generalized statements about algorithms like \siol \ from studying a finite number of \datasetP \ and a small set of group structures. While we conducted a large number of experiments, more work remains to be done to determine which properties of the \datasetP \ are responsible for variations in performance between \datasetP \ and regimes. 

\bibliographystyle{abbrv}
\bibliography{sigproc}

\begin{table}[h]
\centering
{
\footnotesize
\begin{tabular}{cccccc}
Algorithm&Design matrix type&N&d&s&snr \\ \hline
\lasso \ & A & 100 & 1000 & 4 & 0.5\\
\lasso \ & A & 100 & 1000 & 4 & 2\\
\lasso \ & A & 100 & 1000 & 10 & 0.5\\
\lasso \ & A & 100 & 1000 & 10 & 2\\
\lasso \ & A & 1000 & 1000 & 20 & 0.5\\
\lasso \ & A & 1000 & 1000 & 20 & 2\\
\lasso \ & A & 1000 & 1000 & 50 & 0.5\\
\lasso \ & A & 1000 & 1000 & 50 & 2\\
\lasso \ & B & 200 & 1000 & 4 & 0.5\\
\lasso \ & B & 200 & 1000 & 4 & 2\\
\lasso \ & B & 200 & 1000 & 10 & 0.5\\
\lasso \ & B & 200 & 1000 & 10 & 2\\
\lasso \ & B & 1000 & 1000 & 20 & 0.5\\
\lasso \ & B & 1000 & 1000 & 20 & 2\\
\lasso \ & B & 1000 & 1000 & 50 & 0.5\\
\lasso \ & B & 1000 & 1000 & 50 & 2\\
\lasso \ & C & 200 & 1000 & 4 & 0.5\\
\lasso \ & C & 200 & 1000 & 4 & 2\\
\lasso \ & C & 200 & 1000 & 10 & 0.5\\
\lasso \ & C & 200 & 1000 & 10 & 2\\
\lasso \ & C & 1000 & 1000 & 20 & 0.5\\
\lasso \ & C & 1000 & 1000 & 20 & 2\\
\lasso \ & C & 1000 & 1000 & 50 & 0.5\\
\lasso \ & C & 1000 & 1000 & 50 & 2\\
\lasso \ & D & 200 & 100 & 10 & 0.5\\
\lasso \ & D & 200 & 100 & 10 & 2\\
\lasso \ & D & 200 & 100 & 30 & 0.5\\
\lasso \ & D & 200 & 100 & 30 & 2\\
\lasso \ & D & 200 & 1000 & 4 & 0.5\\
\lasso \ & D & 200 & 1000 & 4 & 2\\
\lasso \ & D & 200 & 1000 & 10 & 0.5\\
\lasso \ & D & 200 & 1000 & 10 & 2\\
\lasso \ & D & 1000 & 1000 & 20 & 0.5\\
\lasso \ & D & 1000 & 1000 & 20 & 2\\
\lasso \ & D & 1000 & 1000 & 50 & 0.5\\
\lasso \ & D & 1000 & 1000 & 50 & 2\\
\lasso \ & E & 200 & 200 & 10 & 0.5\\
\lasso \ & E & 200 & 200 & 10 & 2\\
\lasso \ & E & 200 & 200 & 30 & 0.5\\
\lasso \ & E & 200 & 200 & 30 & 2\\
\lasso \ & E & 200 & 1000 & 4 & 0.5\\
\lasso \ & E & 200 & 1000 & 4 & 2\\
\lasso \ & E & 200 & 1000 & 10 & 0.5\\
\lasso \ & E & 200 & 1000 & 10 & 2\\
\lasso \ & E & 1000 & 1000 & 20 & 0.5\\
\lasso \ & E & 1000 & 1000 & 20 & 2\\
\lasso \ & E & 1000 & 1000 & 50 & 0.5\\
\lasso \ & E & 1000 & 1000 & 50 & 2\\
\lasso \ & F & 1000* & 5000* & 20 & 0.5\\
\lasso \ & F & 1000* & 5000* & 20 & 2\\
\lasso \ & F & 1000* & 5000* & 50 & 0.5\\
\lasso \ & F & 1000* & 5000* & 50 & 2\\
\lasso \ & G & 114* & 1260* & 4 & 0.5\\
\lasso \ & G & 114* & 1260* & 4 & 2\\
\lasso \ & G & 114* & 1260* & 10 & 0.5\\
\lasso \ & G & 114* & 1260* & 10 & 2
\end{tabular}
}
\caption{Data configurations for \datasetP \ for the \lasso. Parameters valued marked with a * are intrinsic to the fixed design matrix.}
\label{lassoconfigs}
\end{table}

\begin{table}[h]
\centering
{
\footnotesize
\begin{tabular}{cccccc}
Algorithm&Design matrix type&N&d&s&snr \\ \hline
\grouplasso \ & A & 100 & 1000 & 4 & 0.5\\
\grouplasso \ & A & 100 & 1000 & 4 & 2\\
\grouplasso \ & A & 200 & 1000 & 10 & 0.5\\
\grouplasso \ & A & 200 & 1000 & 10 & 2\\
\grouplasso \ & A & 1000 & 1000 & 10 & 0.5\\
\grouplasso \ & A & 1000 & 1000 & 10 & 2\\
\grouplasso \ & A & 1000 & 1000 & 25 & 0.5\\
\grouplasso \ & A & 1000 & 1000 & 25 & 2\\
\grouplasso \ & B & 200 & 1000 & 4 & 0.5\\
\grouplasso \ & B & 200 & 1000 & 4 & 2\\
\grouplasso \ & B & 200 & 1000 & 10 & 0.5\\
\grouplasso \ & B & 200 & 1000 & 10 & 2\\
\grouplasso \ & B & 1000 & 1000 & 10 & 0.5\\
\grouplasso \ & B & 1000 & 1000 & 10 & 2\\
\grouplasso \ & B & 1000 & 1000 & 25 & 0.5\\
\grouplasso \ & B & 1000 & 1000 & 25 & 2\\
\grouplasso \ & C & 200 & 1000 & 4 & 0.5\\
\grouplasso \ & C & 200 & 1000 & 4 & 2\\
\grouplasso \ & C & 200 & 1000 & 10 & 0.5\\
\grouplasso \ & C & 200 & 1000 & 10 & 2\\
\grouplasso \ & C & 1000 & 1000 & 10 & 0.5\\
\grouplasso \ & C & 1000 & 1000 & 10 & 2\\
\grouplasso \ & C & 1000 & 1000 & 25 & 0.5\\
\grouplasso \ & C & 1000 & 1000 & 25 & 2\\
\grouplasso \ & D & 200 & 100 & 5 & 0.5\\
\grouplasso \ & D & 200 & 100 & 5 & 2\\
\grouplasso \ & D & 200 & 100 & 15 & 0.5\\
\grouplasso \ & D & 200 & 100 & 15 & 2\\
\grouplasso \ & D & 200 & 1000 & 4 & 0.5\\
\grouplasso \ & D & 200 & 1000 & 4 & 2\\
\grouplasso \ & D & 200 & 1000 & 10 & 0.5\\
\grouplasso \ & D & 200 & 1000 & 10 & 2\\
\grouplasso \ & D & 1000 & 1000 & 10 & 0.5\\
\grouplasso \ & D & 1000 & 1000 & 10 & 2\\
\grouplasso \ & D & 1000 & 1000 & 25 & 0.5\\
\grouplasso \ & D & 1000 & 1000 & 25 & 2\\
\grouplasso \ & E & 200 & 200 & 5 & 0.5\\
\grouplasso \ & E & 200 & 200 & 5 & 2\\
\grouplasso \ & E & 200 & 200 & 15 & 0.5\\
\grouplasso \ & E & 200 & 200 & 15 & 2\\
\grouplasso \ & E & 200 & 1000 & 4 & 0.5\\
\grouplasso \ & E & 200 & 1000 & 4 & 2\\
\grouplasso \ & E & 200 & 1000 & 10 & 0.5\\
\grouplasso \ & E & 200 & 1000 & 10 & 2\\
\grouplasso \ & E & 1000 & 1000 & 10 & 0.5\\
\grouplasso \ & E & 1000 & 1000 & 10 & 2\\
\grouplasso \ & E & 1000 & 1000 & 25 & 0.5\\
\grouplasso \ & E & 1000 & 1000 & 25 & 2\\
\grouplasso \ & F & 1000* & 5000* & 10 & 0.5\\
\grouplasso \ & F & 1000* & 5000* & 10 & 2\\
\grouplasso \ & F & 1000* & 5000* & 25 & 0.5\\
\grouplasso \ & F & 1000* & 5000* & 25 & 2\\
\grouplasso \ & G & 114* & 1260* & 2 & 0.5\\
\grouplasso \ & G & 114* & 1260* & 2 & 2\\
\grouplasso \ & G & 114* & 1260* & 5 & 0.5\\
\grouplasso \ & G & 114* & 1260* & 5 & 2
\end{tabular}
}
\caption{Data configurations for \datasetP \ for the \grouplasso. Parameter values marked with a * are intrinsic to the fixed design matrix.}
\label{glconfigs}
\end{table}

\begin{table}[h]
\centering
{
\footnotesize
\begin{tabular}{cccccc}
Algorithm&Design matrix type&N&d&s&snr \\ \hline
\siol \ & A & 500 & 500 & 5 & 0.5\\
\siol \ & A & 500 & 500 & 5 & 2\\
\siol \ & A & 500 & 500 & 10 & 0.5\\
\siol \ & A & 500 & 500 & 10 & 2\\
\siol \ & B & 500 & 500 & 5 & 0.5\\
\siol \ & B & 500 & 500 & 5 & 2\\
\siol \ & B & 500 & 500 & 10 & 0.5\\
\siol \ & B & 500 & 500 & 10 & 2\\
\siol \ & C & 500 & 500 & 5 & 0.5\\
\siol \ & C & 500 & 500 & 5 & 2\\
\siol \ & C & 500 & 500 & 10 & 0.5\\
\siol \ & C & 500 & 500 & 10 & 2\\
\siol \ & D & 500 & 500 & 5 & 0.5\\
\siol \ & D & 500 & 500 & 5 & 2\\
\siol \ & D & 500 & 500 & 10 & 0.5\\
\siol \ & D & 500 & 500 & 10 & 2\\
\siol \ & E & 500 & 500 & 5 & 0.5\\
\siol \ & E & 500 & 500 & 5 & 2\\
\siol \ & E & 500 & 500 & 10 & 0.5\\
\siol \ & E & 500 & 500 & 10 & 2\\
\siol \ & F & 1000* & 5000* & 5 & 0.5\\
\siol \ & F & 1000* & 5000* & 5 & 2\\
\siol \ & F & 1000* & 5000* & 20 & 0.5\\
\siol \ & F & 1000* & 5000* & 20 & 2\\
\siol \ & G & 114* & 1260* & 2 & 0.5\\
\siol \ & G & 114* & 1260* & 2 & 2\\
\end{tabular}
}
\caption{Data configurations for \datasetP \ for \siol. Parameter values marked with a * are intrinsic to the fixed design matrix.}
\label{siolconfigs}
\end{table}

\FloatBarrier
\appendix

\section{Algorithm Details} \label{algodetails}

\subsection{\algoselectPtitle} \label{selectionalgos}

The \algoselectP \ we study are also \algoregP. They are phrased as convex optimization problems over a parameter $\beta$. We minimize the objective until convergence to obtain an optimal parameter value $\hat{\beta}$. Each \component \ of that parameter corresponds to an input-output pair and hence to an element of the \selectionset \ $K$. Hence, we may refer to \componentP \ of $\beta$, \selectionset \ elements and input-output pairs interchangeably. We select an element from $K$ if and only if the corresponding \component \ of $\hat{\beta}$ is non-zero. 

\subsubsection{\lassoF}

The \lasso \ is a classic, well-studied algorithm for discovering associations between input variables and a single output variable. Hence, $t=1$ and $K$ and the \componentP  \  of $\beta$ correspond to the set of input variables. \lassoF \ uses least-squares regression coupled with an $L_1$-penalty term on the \componentP   \ of $\beta$. This causes shrinkage which in turn leads to \componentP \ corresponding to input variables with little or no association to the output to become exactly zero. The formula is given below. $\beta$ is a $d$ by $1$ vector. 

\begin{equation*}
\min_\beta \frac{1}{2N} ||y - X\beta||_2^2 + \lambda ||\beta||_1
\end{equation*}

%Meinshausen and B{\"u}hlmann \cite{ss} chose this algorithm to demonstrate the benefits of \stabilityselection. They showed that \stabilityselection \ increases the \accuracy \ of \lasso \ and provides accurate error control. 

\subsubsection{\grouplassoF}

The \grouplasso \ is an extension of the Lasso. It is applied in cases where, instead of selecting individual input variables, we wish to select groups of input variables. Each input variable must belong to exactly one group, and $t=1$ holds. In $\hat{\beta}$, either all \componentP  \  belonging to the same group are equal to zero or all are non-zero. The group-wise selection property is achieved by imposing an $L_2$-norm penalty on each group of \componentP. This penalty is called a {\it group penalty}. 

\begin{equation*}
\min_\beta \frac{1}{2N} ||y - X\beta||_2^2 + \lambda \sum_{g \in G} \theta_g ||\beta_g||_2
\end{equation*}

$\beta$ is a $d$ by $1$ vector. $G$ is the set of groups - a set of subsets of $\{1,..,d\}$ such that all $g \in G$ are disjoint and their union is $\{1,..,d\}$. $\beta_g$ is the sub-vector of $\beta$ containing the \componentP   \  of $\beta$ in $g$. This algorithm has additional \tuningP \ $\theta_g$ that allow us to weight the penalties applied to different groups. Note that this algorithm reduces to the \lasso \ if every input variable belongs to its own group of size one and the $\theta_g$ are all equal to one.

\subsubsection{\siolpureF} \label{siol}

\siolfirstF \ extends \grouplasso \ in several respects. Firstly, it admits multiple output variables. This means that $\beta$ becomes a $d$ by $t$ matrix. Secondly, it admits group penalties across \componentP \ of $\beta$ corresponding to the same output variable. Thirdly, it admits overlapping groups. Fourthly, it combines $L_2$-based group penalties with single-\component \ $L_1$-penalties. This allows the algorithm to select isolated elements while encouraging the selection of groups as well as blocks of input and output groups which have a high overlap. 

%\begin{gather*}
%\min_\beta \frac{1}{2N} ||Y - X\beta||_2^2 + \lambda(w\sum_{i=1}^d\sum_{j=1}^t |\beta_{ij}|_1 + \\
%\sum_{g \in G}\sum_{j=1}^t \theta_g ||\beta_{gj}||_2  + \sum_{i=1}^d\sum_{h \in H} \phi_h||\beta_{ih}||_2) \text{,}\\
%\text{$g$ and $h$ may mutually overlap}
%\end{gather*}

\begin{gather*}
\min_\beta \frac{1}{2N} ||Y - X\beta||_2^2 + \lambda(w\sum_{i=1}^d\sum_{j=1}^t |\beta_{ij}| + \sum_{g \in G}\sum_{j=1}^t \theta_g ||\beta_{gj}||_2  + \sum_{i=1}^d\sum_{h \in H} \phi_h||\beta_{ih}||_2)
\end{gather*}

$G$ is the set of input groups as before except that input groups do not have to be disjoint and not all input variables have to be in an input group. $\beta_{gj}$ is the sub-vector of matrix $\beta$ of \componentP \ corresponding to output variable $j$ and input variables in $g$. $H$ / $\beta_{ih}$ are the equivalent to $G$ / $\beta_{gj}$ for output groups. $w$, $\theta_g$ and $\phi_h$ allow us to weight different penalty terms. Note that if no output groups are chosen, this objective function decomposes over output variables into independent optimization problems. If also no input groups are chosen, the algorithm reduces to the Lasso.

\section{Experimental details} \label{expdetails}

\paragraph{\grouplassoF} We only consider groups of equal size and set the group weights $\theta_g$ to 1.

\paragraph{\siolF} We set $\theta_g$ / $\theta_h$ to be equal to the square root of the size of the corresponding group. This is a popular choice. We run \siol \ both with $w=0.5$ and $w=2$. Note that instead of fixing tuning parameters such as the $\theta_g$ / $\theta_h$ or $w$ in advance, we can use the error control schemes to set some or all of them automatically. This is beyond the scope of this paper. We only tune the parameter $\lambda$ (which represents the overall amount of regularization) along with, for \stabilityselection \ and \screenandclean, the selection threshold $\pi$. 

\paragraph{\xvalF} \ We compute the \xval \ error for a sequence of values $\lambda_1, \lambda_2, .., \lambda_L$, where $\lambda_1$ is large enough so that the initial selection on the training set is empty. We set $\lambda_l = 0.98*\lambda_{l-1}$ for $2 \le l \le L$ and we set $L$ to be very large, so that we consider every sensible choice of $\lambda$. A popular rule of thumb is that \xval \ becomes unreliable once there is an output variable for which the number of selected input variables exceeds one half the number of data points in the training set. This is the ``cutoff'' value for $L$ that we use in this paper. We use ten folds ($k = 10$). 

\paragraph{Stability Selection} We follow the example of \cite{ss}. We run 100 iterations ($I=100$) and sub-sample one half of the \dataset \ at each iteration ($p = \frac{1}{2}$). %, and run the \subalgo \ with $\lambda$-max (see section \ref{lmax}). 
%We set $\lambda_1$ in $\Lambda$ to be large enough so that the selection will be empty and set $\lambda_l = 0.98*\lambda_{l-1}$ for $2 \le l \le L$. We tune $L$ (and hence $\lambda_{min}$) so that $\hat{q} \approx \sqrt{0.8*|K|}$ as discussed in section \ref{sserror}. While $\lambda$-max could be useful for \xval \ and \screenandclean, because it is traditionally not combined with those methods, we only use $\lambda$-max with \stabilityselection.
In many experiments, we use the heuristic of setting $\lambda$ to achieve $\hat{q} \approx \sqrt{0.8*|K|}$. As $\lambda$ decreases, $\hat{q}$ almost certainly increases. Thus, there is usually a unique value or a small range of values of $\lambda$ that exactly achieves $\hat{q} = \sqrt{0.8*|K|}$. In practice, we consider the same sequence of $\lambda$-values as in \xval, and then choose the smallest value of $\lambda$ in that sequence for which $\hat{q} > \sqrt{0.8*|K|}$ holds. 

{\bf Screen and Clean} We use 10 top-level data splits ($I = 10$). Within each screen phases, we use 10-fold cross-validation as described above.

\section{Running time} \label{runtime}

In this section, we compare the running time of different error control schemes. For each of \stabilityselection, \screenandclean \ and \xval, the running time is almost completely determined by the number of times they execute the sub-algorithm. In our experiments, we always observed that the running time was close to the value obtained by multiplying the time it took to run the sub-algorithm once with the number of repetitions. \stabilityselectionF \ executes the sub-algorithm $I$ times, there $I$ is the number of iterations. \screenandcleanF \ executes the sub-algorithm $Ik$ times, where $I$ is the number of separate screen / clean phases and $k$ is the number of folds used for the \xval \ sub-routine. \xvalF \ executes the sub-algorithm $k$ times. 

One minor point is that \xval \ runs the sub-algorithm on almost the entire \dataset, whereas \stabilityselection \ and \screenandclean \ use only roughly $\frac{1}{2}$ of the \dataset \ in each iteration. Hence, \xval \ may take a little longer on each iteration. This assumes that we set the sub-sampling parameter $p$ to $\frac{1}{2}$ for \stabilityselection. In our experiments, we always did this (consistent with \cite{ss}) but we may get away with a smaller value to achieve a speed-up.

Also, note that if we set $\lambda$ to achieve $\hat{q} \approx \sqrt{0.8*|K|}$ for \stabilityselection, then we may not have to run the sub-algorithm for small values of $\lambda$ as we have to do for \xval \ and \screenandclean. This may cause a significant speed-up, as it is a well-known fact that \lasso \ and related algorithms converge much more slowly when the value of $\lambda$ is small. In our experiments, we used the same values of $\lambda$ for all error control schemes and did not analyze this further.

In our experiments, we chose $k=10$ and $I=10$ for \screenandclean \ and $I=100$ for \stabilityselection. Hence, \screenandclean \ and \stabilityselection \ had almost the same running time. These values are consistent with other literature.

The benefit of increasing $I$ and $k$ lies in eliminating random noise obtained from random data splitting. Of course, this means that growing $I$ and $k$ yields diminishing returns. We cannot increase $I$ and $k$ arbitrarily and expect our results to keep improving. We did not investigate this trade-off.

Finally, we note that all runs of the \subalgo \ can be trivially parallelized.

\section{Detailed Results} \label{detailedresults}

\begin{figure*}[!ht]
\centering
\includegraphics[scale=2]{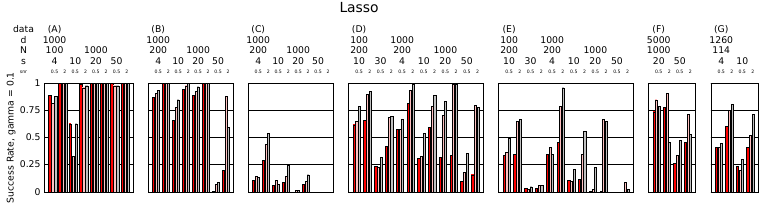}
\includegraphics[scale=2]{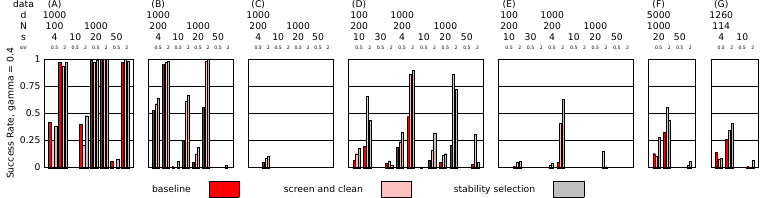}
\caption{Probability (over 100 random \datasetP) that there exists a \tuning \ value such that at least $\lceil \gamma*s \rceil$ positives and zero negatives are selected. Parameter values for data configuration are shown above the charts. Results shown pertain to \lasso.}
\label{lassobars}
\end{figure*}

\begin{figure*}[!ht]
\centering
\includegraphics[scale=2]{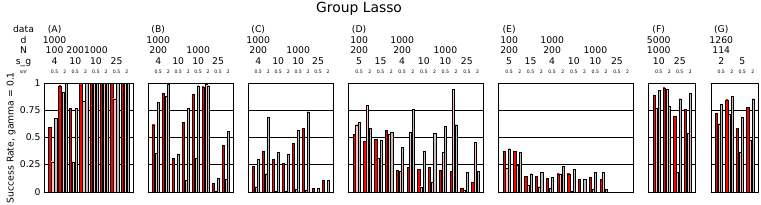}
\includegraphics[scale=2]{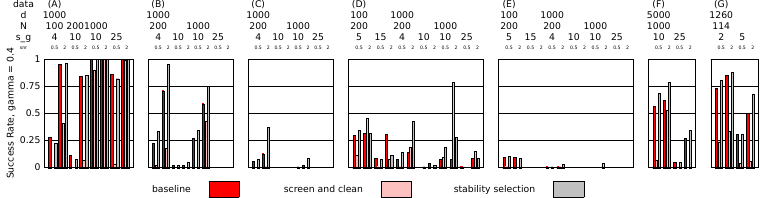}
\caption{Probability (over 100 random \datasetP) that there exists a \tuning \ value such that at least $\lceil \gamma*s \rceil$ positives and zero negatives are selected. Parameter values for data configuration are shown above the charts. Results shown pertain to \grouplasso.}
\label{glbars}
\end{figure*}

\begin{figure*}[!ht]
\centering
\includegraphics[scale=2]{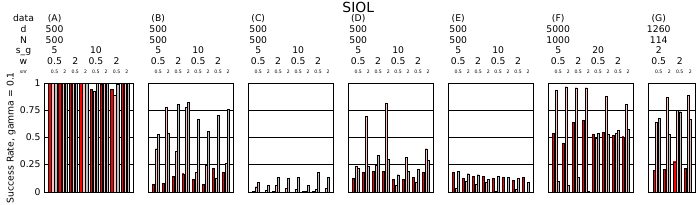}
\includegraphics[scale=2]{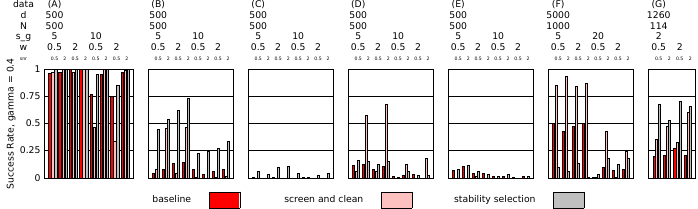}
\caption{Probability (over 100 random \datasetP) that there exists a \tuning \ value such that at least $\lceil \gamma*s \rceil$ positives and zero negatives are selected. Parameter values for data configuration are shown above the charts. Results shown pertain to \siol.}
\label{siolbars}
\end{figure*}

In figures \ref{lassobars}, \ref{glbars}, and \ref{siolbars}, we show by-configuration results for the statistical power of the regimes. We focus on the important case $\fpr = 0$ and show the largest possible number of true positives that can be selected. To achieve a fair comparison, for each regime, we maximize $\tpr$ over a single \tuning. For \stabilityselection, we set $\lambda$ to achieve $\hat{q} \approx \sqrt{0.8*|K|}$ and maximize over $\pi$. For \screenandclean, we maximize over $\pi$. For the baseline regime, we maximize over $\lambda$. In the charts, we show the probability that the number of positives we can select without selecting a negative is $\lceil\gamma* s\rceil$ or larger ($s = |S^*|$), for $\gamma = 0.1$ and $\gamma = 0.4$. Probabilities are with respect to randomly drawn \datasetP \ within each data configuration and are estimated as empirical averages over the 100 \datasetP \ we generated for each configuration. In figure \ref{siolbars}, we break down results also by the value of $w$, the \tuning \ in \siol.

\end{document}